\documentclass[runningheads]{llncs}
\usepackage[T1]{fontenc}
\usepackage{graphicx,verbatim}
\usepackage{multirow}
\usepackage[table,xcdraw]{xcolor}
\usepackage[normalem]{ulem}
\useunder{\uline}{\ul}{}
\usepackage{booktabs}
\usepackage[colorlinks=true, linkcolor=blue, citecolor=black, urlcolor=magenta]{hyperref}
\usepackage{amsfonts}
\usepackage{amsmath} 
\definecolor{seencolor}{RGB}{212, 233, 201}
\definecolor{unseencolor}{RGB}{255, 235, 171}
\usepackage{marvosym}

\definecolor{seentextcolor}{RGB}{112, 173, 71}
\newcommand{\seentxt}[1]{\textbf{\textit{\color{seentextcolor}{#1}}}}
\definecolor{unseentextcolor}{RGB}{249, 187, 0}
\newcommand{\unseentxt}[1]{\textbf{\textit{\color{unseentextcolor}{#1}}}}

\begin{document}
\title{BenchReAD: A systematic benchmark \\ for retinal anomaly detection}

\author{
Chenyu Lian\inst{1} \and
Hong-Yu Zhou\inst{2}\textsuperscript{(\Letter)} \and
Zhanli Hu\inst{3} \and
Jing Qin\inst{1}\textsuperscript{(\Letter)}
}

%
\authorrunning{C. Lian et al.}
%
\institute{The Center for Smart Health, School of
Nursing, the Hong Kong Polytechnic University, Hong Kong, China \\
\email{chenyu.lian@connect.polyu.hk, harry.qin@polyu.edu.hk} \and
School of Biomedical Engineering, Tsinghua University, Beijing, China \\
\email{whuzhouhongyu@gmail.com} \and
Research Center for Medical AI, Shenzhen Institute of Advanced Technology, Chinese Academy of Sciences, Shenzhen, China. \\
\email{zl.hu@siat.ac.cn}
}

\maketitle              
\begin{abstract}
Retinal anomaly detection plays a pivotal role in screening ocular and systemic diseases.
Despite its significance, progress in the field has been hindered by the absence of a comprehensive and publicly available benchmark, which is essential for the fair evaluation and advancement of methodologies.
Due to this limitation, previous anomaly detection work related to retinal images has been constrained by (1) a limited and overly simplistic set of anomaly types, (2) test sets that are nearly saturated, and (3) a lack of generalization evaluation, resulting in less convincing experimental setups.
Furthermore, existing benchmarks in medical anomaly detection predominantly focus on one-class supervised approaches (training only with negative samples), overlooking the vast amounts of labeled abnormal data and unlabeled data that are commonly available in clinical practice.
To bridge these gaps, we introduce a benchmark for retinal anomaly detection, which is comprehensive and systematic in terms of data and algorithm.
Through categorizing and benchmarking previous methods, we find that a fully supervised approach leveraging disentangled representations of abnormalities (DRA) achieves the best performance but suffers from significant drops in performance when encountering certain unseen anomalies.
Inspired by the memory bank mechanisms in one-class supervised learning, we propose NFM-DRA, which integrates DRA with a Normal Feature Memory to mitigate the performance degradation, establishing a new SOTA.
The benchmark is publicly available at \url{https://github.com/DopamineLcy/BenchReAD}.

\keywords{Benchmark  \and anomaly detection \and retinal imaging.}

\end{abstract}

\section{Introduction}
Retinal imaging is critical in screening a variety of ocular and systemic conditions, such as diabetic retinopathy, glaucoma, and myopia~\cite{han2021application,li2021association,narasimha2006robust}.
Despite the widespread application of anomaly detection in medical imaging, the field still lacks a \textbf{comprehensive and systematic benchmark} to standardize evaluation protocols and facilitate the development of novel methods.
Existing benchmarks of medical anomaly detection related to retinal images often rely on \textbf{single datasets} that primarily focus on \textbf{one or a few diseases}~\cite{aptos2019blindness,li2019attention,kermany2018identifying,hu2019automated} for both model development and evaluation, which fails to provide a robust and generalizable assessment~\cite{bao2024bmad,cai2024medianomaly}.
We also notice that the widely used OCT 2017 dataset~\cite{kermany2018identifying,zhai2024dual,han2021gan,guo2024recontrast} has become unsuitable as a standalone benchmark because of \textbf{performance saturation}.
Furthermore, most existing benchmarks focus on one-class supervised methods that only adopt negative samples for training, which originate from industrial anomaly detection where abnormal instances are rare~\cite{liu2023simplenet,roth2022towards,li2021cutpaste,deng2022anomaly}.
In contrast, retinal imaging datasets often contain many disease samples, necessitating a more nuanced evaluation framework.

\begin{figure}[!t]
    \centering
    \includegraphics[width=\linewidth]{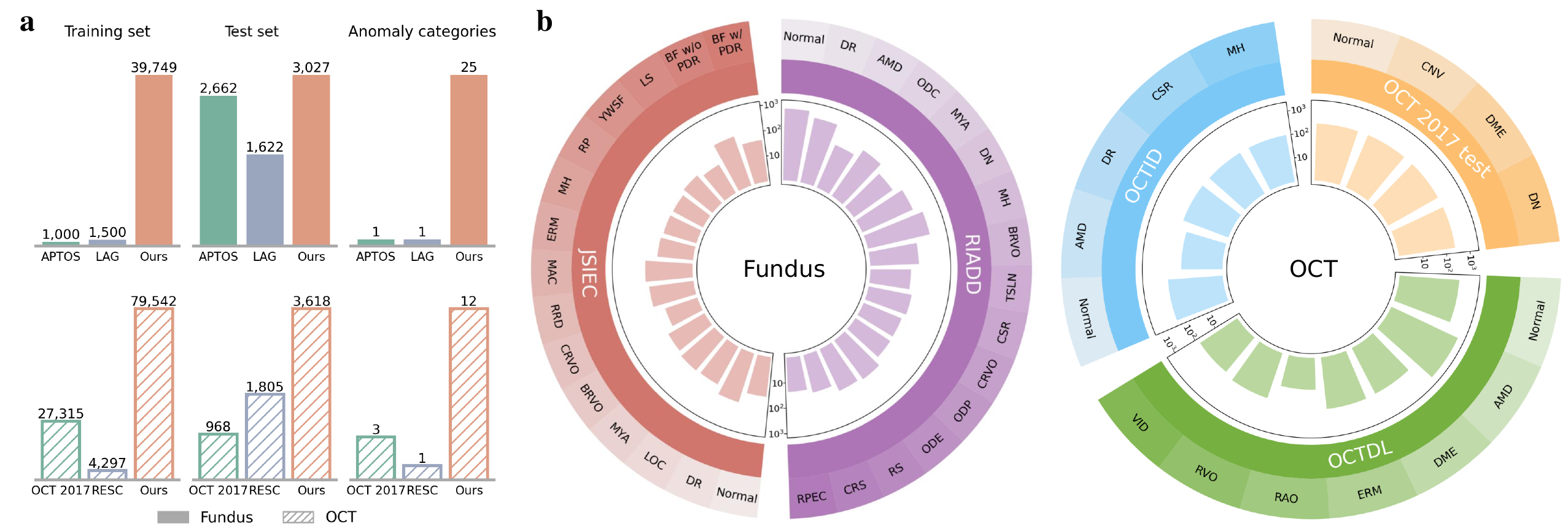}
    \caption{(a) Comparison among widely used datasets for retinal anomaly detection (APTOS~\cite{aptos2019blindness}, LAG~\cite{li2019attention}, OCT 2017~\cite{kermany2018identifying}, and RESC~\cite{hu2019automated}) and our BenchReAD.
    (b) Overview of the test datasets included in the proposed benchmark.}
    \label{fig:metadata}
\end{figure}
To address these limitations, we introduce BenchReAD, a systematic benchmark for retinal anomaly detection that encompasses two of the most widely used imaging modalities: fundus photography and optical coherence tomography (OCT).
Our benchmark is designed to be systematic and comprehensive in two key dimensions: dataset and methodologies.
From the \textbf{dataset} perspective, as illustrated in Figure~\ref{fig:metadata}, BenchReAD incorporates notably larger datasets with a more diverse range of anomaly categories compared to existing benchmarks, ensuring a rigorous assessment of \textbf{generalizability}.
From the \textbf{methodology} perspective, we systematically categorize anomaly detection approaches into four groups based on their supervision levels: unsupervised, one-class supervised, semi-supervised, and fully supervised methods.
Through extensive benchmarking, we find that a fully supervised approach leveraging disentangled representations of abnormalities (DRA) achieves the highest performance. However, it suffers from a significant drop in effectiveness when encountering certain unseen anomalies.
Inspired by the memory banks in one-class supervised learning, we propose \textbf{NFM-DRA} by integrating DRA with a \uline{N}ormal \uline{F}eature \uline{M}emory to mitigate the performance drops, leading to a new SOTA.
\section{Benchmark}
\subsection{Datasets}
\label{sec:datasets}
The benchmark includes two of the most widely used retinal imaging modalities: fundus photography and optical coherence tomography (OCT).
The test sets contain both \seentxt{seen} (present in the training set) and \unseentxt{unseen} (absent from the training set) anomalies.

\noindent\textbf{The fundus benchmark} is constructed from four publicly available datasets.
The training and validation sets are derived from EDDFS~\cite{XIA2024117151} and BRSET~\cite{nakayama2024brset}, while the test set is composed of the RIADD~\cite{pachade2021retinal} and JSIEC~\cite{cen2021automatic}.
We retain four common retinal anomalies of diabetic retinopathy (DR), age-related macular degeneration (AMD), glaucoma, and myopia (MYA) in these two datasets.
After preprocessing and data structuring, the training set comprises 23,361 normal images and 16,388 abnormal images, with an additional 100 normal and 100 abnormal images reserved for validation.
Figure~\ref{fig:metadata} (b) illustrates distribution of test sets categories, ensuring statistical robustness by excluding anomaly categories with fewer than 20 images.
The RIADD dataset includes 669 normal images and 1,634 abnormal images, covering both \seentxt{seen} anomalies (DR, AMD, optic disc cupping (ODC, highly associated with glaucoma), MYA, and drusens (DN)) and \unseentxt{unseen} anomalies (macular hole (MH), branch retinal vein occlusion (BRVO), tractional retinal detachment (TSLN), central serous retinopathy (CSR), central retinal vein occlusion (CRVO), optic disc pallor (ODP), optic disc edema (ODE), retinoschisis (RS), choroidal rupture syndrome (CRS), and retinal pigment epithelium changes (RPEC)).
The JSIEC dataset comprises 38 normal images and 686 abnormal images, featuring \seentxt{seen} categories of DR, large optic cup (LOC, highly associated with glaucoma), MYA as well as \unseentxt{unseen} categories including BRVO, CRVO, rhegmatogenous RD (RD), maculopathy (MAX), epiretinal membrane (ERM), MH, Retinitis pigmentosa (RP), yellow-white spots-flecks (YWSF), laser Spots (LS), blur fundus without PDR (BF w/o PDR), and blur fundus with suspected PDR (BF w/ PDR).

\noindent\textbf{The OCT benchmark} is established using three public datasets: OCT 2017~\cite{kermany2018identifying}, OCTDL~\cite{kulyabin2024octdl}, and OCTID~\cite{gholami2020octid}.
The training and validation sets are derived from OCT 2017, a large-scale classification dataset containing normal images and three abnormal classes: choroidal
neovascularization (CNV), diabetic macular edema (DME), and drusen deposits (DN).
The training set includes 25,840 normal and 53,702 abnormal images, while the validation set includes 50 images per class (200 images in total).
The test sets include the OCT 2017 test set, OCTDL, and OCTID, as shown in Figure~\ref{fig:metadata} (b).
The OCT 2017 test set includes 250 images per category (normal, CNV, DME, and DN).
OCTDL includes 332 normal and 1,732 abnormal images, with two \seentxt{seen} anomalies (AMD and DME) and four \unseentxt{unseen} anomalies: (epiretinal membrane (ERM), RAO, retinal vein occlusion (RVO), and vitreomacular interface disorders (VID)).
OCTID consists of 206 normal and 336 abnormal images, including two \seentxt{seen} anomalies (AMD and DR), and two \unseentxt{unseen} anomalies: central serous retinopathy (CSR) and MH.
\subsection{Data and algorithms across different supervision levels}
\label{sec:supervision_levels}
\textbf{Supervision level settings in the training set.}
To systematically evaluate the performance of anomaly detection methods under different supervision levels, we partition the training set into labeled and unlabeled subsets.
Specifically, the labeled subset constitutes one-third of the training data, while the unlabeled subset accounts for the remaining two-thirds.
Formally, the labeled normal image subset is denoted as $\mathcal{N}_x$, with corresponding labels $\mathcal{N}_y$; the labeled abnormal image subset is denoted as $\mathcal{A}_x$, with corresponding labels $\mathcal{A}_y$; and the unlabeled image subset $\mathcal{U}_x$ contains images without any associated labels.

\noindent\textbf{Algorithms across different supervision levels.}
Unlike previous benchmarks for medical anomaly detection that predominantly focus on one-class supervised algorithms, our benchmark systematically categorizes approaches into four groups based on supervision levels:
\textit{\textbf{Unsupervised}} methods require no annotations and make use of all available data ($\mathcal{N}_x$, $\mathcal{A}_x$, and $\mathcal{U}_x$).
We choose SoftPatch~\cite{jiang2022softpatch} (NeurIPS 2023) as the representative method.
\textit{\textbf{One-class supervised}} methods rely exclusively on normal images and implicitly use their corresponding labels ($\mathcal{N}_x$, $\mathcal{N}_y$). They are trained to model the distribution of normal data and detect deviations as anomalies.
We select three variants: EDC~\cite{guo2023encoder} (distillation-based, TMI 2023), SimpleNet~\cite{liu2023simplenet} (self-supervised, CVPR 2023), and PatchCore~\cite{roth2022towards} (memory bank-based, CVPR 2022).
\textit{\textbf{Semi-supervised}} methods represented by DDAD-ASR~\cite{cai2023dual} (MedIA 2023) extend one-class supervised methods by incorporating additional unlabeled data $\mathcal{U}_x$ alongside $\mathcal{N}_x$ and $\mathcal{N}_y$ to enhance their performance.
\textit{\textbf{Fully supervised}} methods utilize both labeled normal and abnormal images along with their labels ($\mathcal{N}_x$, $\mathcal{N}_y$, $\mathcal{A}_x$, $\mathcal{A}_y$) and aims to detect both seen and unseen anomalies~\cite{ding2022catching}.
DRA~\cite{ding2022catching} (CVPR 2022), an open-set supervised anomaly detection method, is benchmarked here.
\subsection{The proposed NFM-DRA with normal feature memory}
\label{sec:NFM}
The fully supervised method DRA~\cite{ding2022catching} is observed to achieve the best performance, as we will discuss in Section~\ref{sec:results_and_discussions}.
However, since it is trained on both normal and abnormal samples, the predictions tend to rely on capturing previously seen abnormal features instead of similarities to normal features.
This reliance on seen anomaly features reduces robustness when encountering certain unseen anomalies.
To address this limitation, we draw inspiration from the memory bank mechanism used in anomaly detection~\cite{gong2019memorizing,roth2022towards} and introduce a simple yet effective improvement.
Specifically, we construct a normal feature memory $\mathcal{M}$ to store features of normal samples.
By comparing test image features $\mathcal{X}$ against this memory, we get representative feature $\boldsymbol{x}^*$ from $\mathcal{X}$ and its closest feature $\boldsymbol{m}^*$ from $\mathcal{M}$:
\begin{equation}
\begin{aligned}
\boldsymbol{x}^*, \boldsymbol{m}^* &= \underset{\boldsymbol{x} \in \mathcal{X}}{\arg \max}\ \underset{\boldsymbol{m} \in \mathcal{M}}{\arg \min} \left\| \boldsymbol{x} - \boldsymbol{m} \right\|_2. \\
\end{aligned}
\end{equation}
Then, anomaly scores can be computed through:
\begin{equation}
\label{eq:g}
\begin{aligned}
g\left(\mathcal{X}; {\mathcal{M}}\right)=\left(1-\frac{\exp\left(\left\|\boldsymbol{x}^*-\boldsymbol{m}^*\right\|_2\right)}{\sum_{\boldsymbol{m} \in \mathcal{N}^{\mathcal{M}}_b\left(\boldsymbol{m}^*\right)} \exp \left\|\boldsymbol{x}^{*}-\boldsymbol{m}\right\|_2}\right) \cdot \left\|\boldsymbol{x}^*-\boldsymbol{m}^*\right\|_2,
\end{aligned}
\end{equation}
where $\mathcal{N}^{\mathcal{M}}_b\left(\boldsymbol{m}^*\right)$ denotes $b$ features in $\mathcal{M}$ that are nearest to $\boldsymbol{m}^*$.
We refine the prediction scores of DRA using anomaly values derived from Equation~\ref{eq:g}, generating the final anomaly scores:
\begin{equation}
\label{eq:report_representation_obtain}
f\left(\mathcal{X}; {\mathcal{M}, {\rm DRA}}\right)=
\frac{1}{2}
\left[
g\left(\mathcal{X}; {\mathcal{M}}\right) + {\rm DRA}\left(\mathcal{X}\right)
\right].
\end{equation}

\begin{figure}[!t]
    \centering
    \includegraphics[width=1\linewidth]{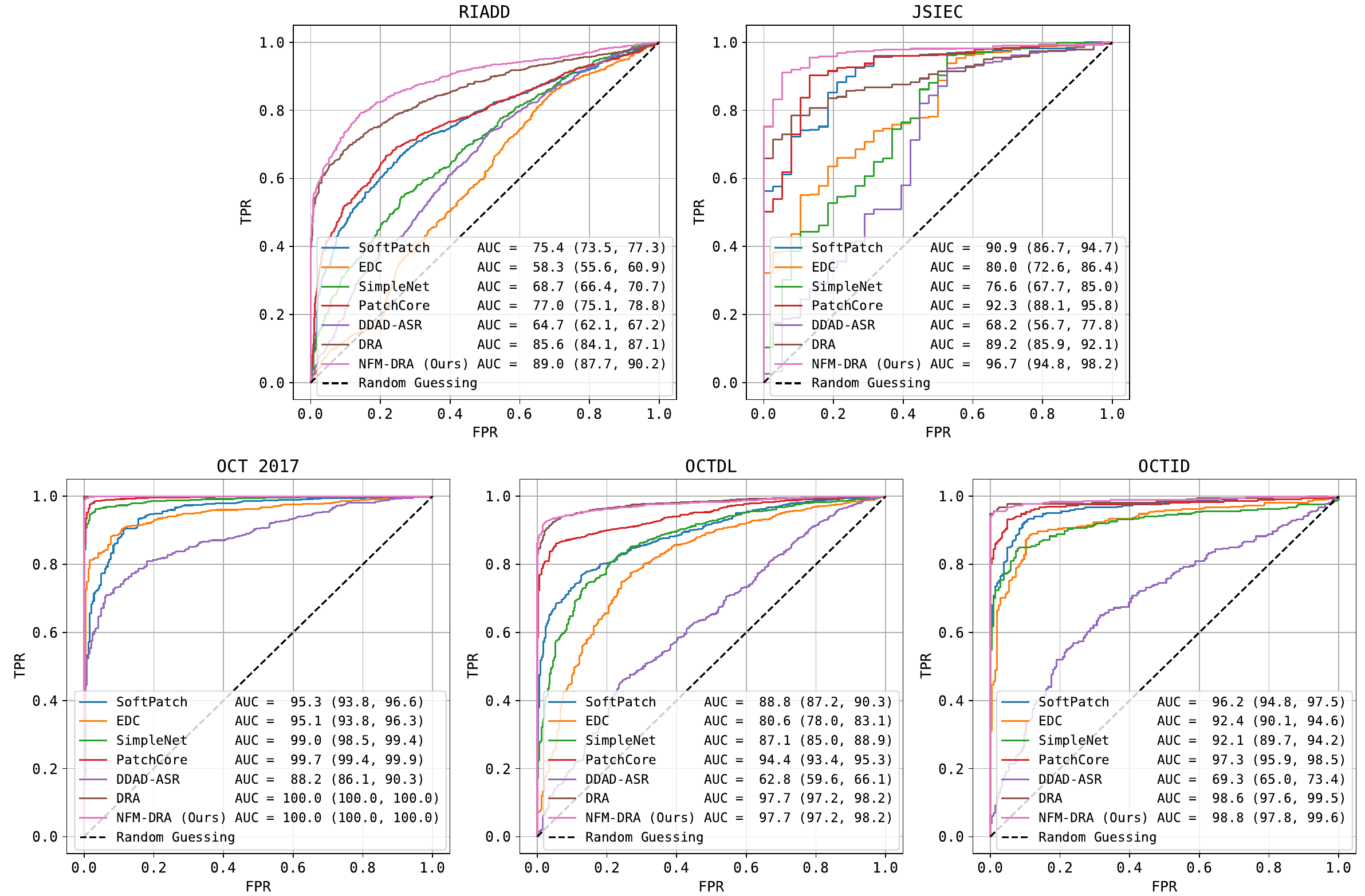}
    \caption{ROC curves of distinguishing normal samples against all abnormal ones on test sets.
    Corresponding AUCs (\%) are marked alongside 95\% confidence intervals.}
    \label{fig:All_Datasets_AUROC}
\end{figure}
\section{Experiments and analyses}
\subsection{Results and discussions}
\label{sec:results_and_discussions}
\textbf{Overall evaluation of distinguishing normal and abnormal samples.}
Figure~\ref{fig:All_Datasets_AUROC} presents the comparison of ROC curves and their corresponding AUCs with 95\% confidence intervals (CIs).
The threshold-independent metric measures the ability of the model to distinguish between normal and abnormal retinal images across all possible decision thresholds.
The fully supervised method, DRA, outperforms other benchmarked approaches across four out of the five test sets, achieving notably high AUC values with narrow confidence intervals.
PatchCore, a one-class supervised method, ranks second in overall performance and attains the highest AUC score on the JSIEC dataset. 
While other methods exhibit competitive performance on specific datasets, they generally lag behind DRA and PatchCore.
The proposed NFM-DRA is the best-performing approach, showing notable improvements, particularly on the RIADD and JSIEC datasets.

\begin{figure}[!t]
    \centering
    \includegraphics[width=\linewidth]{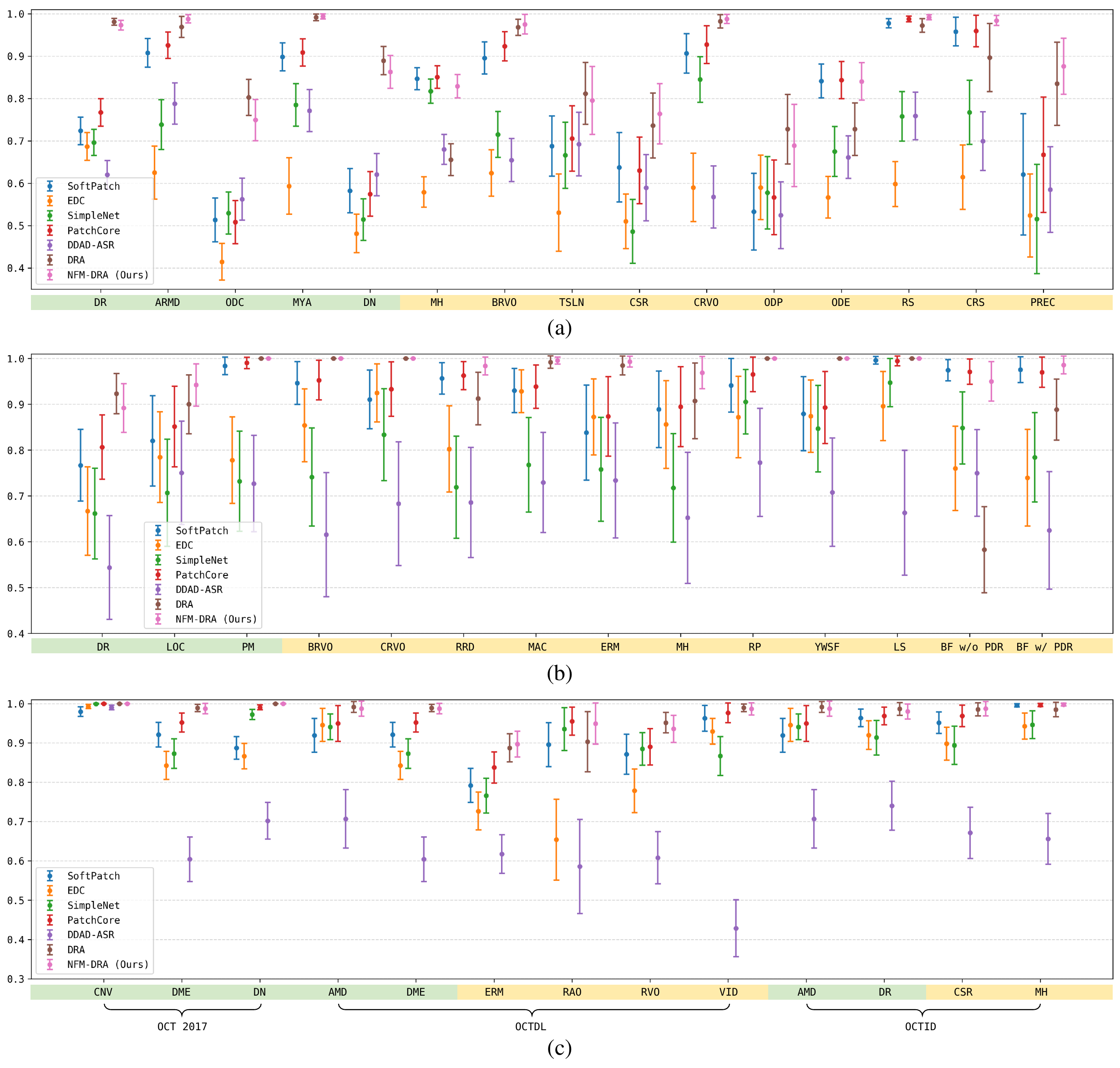}
    \caption{AUROCs with 95\% confidence intervals for normal samples compared to each abnormal category on the (a) RIADD and (b) JSIEC datasets for fundus benchmarking, as well as (c) OCT test datasets.
    \colorbox{seencolor}{Seen} and \colorbox{unseencolor}{unseen} annotations highlight whether the anomaly categories appear in the training set or not, respectively.
    }
    \label{fig:AUROC_abc}
\end{figure}
\noindent\textbf{Fully supervised method DRA achieves the best performance, while one-class supervised method PatchCore exhibits greater robustness.}
In addition to the overview comparisons (Figure~\ref{fig:All_Datasets_AUROC}), AUROCs with 95\% CIs for normal samples compared to each abnormal category in the RIADD, JSIEC and OCT test sets are presented in Figure~\ref{fig:AUROC_abc} (a), (b), and (c), respectively.
DRA consistently outperforms other methods on seen anomalies and also performs well on unseen anomalies.
However, for certain anomaly categories such as MH, ODE, RS, and CRS in the RIADD dataset, the performance is notably weaker, and PatchScore achieves the best performance.
A similar trend is observed in the JSIEC dataset, where DRA consistently surpasses counterparts on seen categories but underperforms on specific unseen anomalies, including CRVO, BF w/o PDR, and BF w/ PDR.
Notably, DRA performs the worst among all comparative methods on the BF w/o PDR category.
The observed contradiction between the strong overall performance of supervised methods and their unstable performance on unseen anomalies underscores the need to improve robustness.

\noindent\textbf{The proposed NFM-DRA achieves a new SOTA and shows robustness for unseen anomalies.}
Motivated by the findings above, we propose NFM-DRA to boost the robustness of DRA across diverse anomaly types by leveraging a \uline{N}ormal \uline{F}eature \uline{M}emory.
Specifically, we augment prediction scores of DRA with anomaly scores derived from comparing the testing images with the proposed normal feature memory, thereby obtaining the final anomaly scores (refer to Section~\ref{sec:NFM} for more details).
As illustrated in Figure~\ref{fig:All_Datasets_AUROC}, the proposed NFM-DRA is the best method on fundus datasets and delivers performance comparable to DRA on OCT datasets, further enhancing robustness.
NFM-DRA mitigates the significant performance drops in detecting unseen anomalies such as MH and ODE in the RIADD dataset, as well as BF w/o PDR in the JSIEC dataset, as shown in Figure~\ref{fig:AUROC_abc} (a) and (b).
Furthermore, the threshold-dependent performance is improved, as evidenced by the average F1 scores in Table~\ref{tab: RIADD} and~\ref{tab: OCT}.

%
\begin{table}[!tp]
\caption{Threshold-dependent results (\%) on fundus test sets, presenting F1 scores, specificity and sensitivity.
Categories of \colorbox{seencolor}{seen} and \colorbox{unseencolor}{unseen} anomalies are highlighted to indicate whether they are included in the training set.
The \textbf{best} results are shown in bold, and the \uline{second-best} ones are underlined.
}
\label{tab: RIADD}
\centering
\resizebox{\linewidth}{!}{
\begin{tabular}{c|c|ccc|ccc|ccc|ccc|ccc|ccc|ccc}
\toprule
                          &                                    & \multicolumn{3}{c|}{SoftPatch}      & \multicolumn{3}{c|}{EDC}              & \multicolumn{3}{c|}{SimpleNet}         & \multicolumn{3}{c|}{PatchCore}         & \multicolumn{3}{c|}{DDAD-ASR}          & \multicolumn{3}{c|}{DRA}                     & \multicolumn{3}{c}{NFM-DRA (Ours)}                      \\ \cline{3-23} 
\multirow{-2}{*}{Dataset} & \multirow{-2}{*}{Normal   vs}      & F1         & SPC  & SEN            & F1            & SPC & SEN            & F1            & SPC  & SEN            & F1            & SPC  & SEN            & F1            & SPC  & SEN            & F1            & SPC        & SEN            & F1             & SPC            & SEN            \\ \hline
                          & All abnormal                       & 82.7       & 5.8  & 97.7           & \textbf{83.0} & 0.0 & \textbf{100.0} & \textbf{83.0} & 0.9  & {\ul 99.8}     & 82.4          & 13.6 & 94.9           & 37.4          & 88.6 & 24.1           & 79.3          & {\ul 90.3} & 68.3           & 70.0           & \textbf{99.3}  & 54.0           \\
                          & \cellcolor[HTML]{E2EFDA}DR         & 54.8       & 5.8  & 97.0           & 54.5          & 0.0 & \textbf{100.0} & 54.6          & 0.9  & {\ul 99.8}     & 56.1          & 13.6 & 95.3           & 26.5          & 88.6 & 18.2           & {\ul 90.0}    & {\ul 90.3} & 95.0           & \textbf{93.2}  & \textbf{99.3}  & 88.3           \\
                          & \cellcolor[HTML]{E2EFDA}AMD       & 18.0       & 5.8  & \textbf{100.0} & 17.1          & 0.0 & \textbf{100.0} & 17.2          & 0.9  & \textbf{100.0} & 19.3          & 13.6 & \textbf{100.0} & 33.3          & 88.6 & 42.0           & {\ul 64.6}    & {\ul 90.3} & 92.8           & \textbf{87.9}  & \textbf{99.3}  & 84.1           \\
                          & \cellcolor[HTML]{E2EFDA}ODC        & 30.8       & 5.8  & 92.3           & {\ul 31.7}    & 0.0 & \textbf{100.0} & {\ul 31.7}    & 0.9  & {\ul 99.4}     & 29.7          & 13.6 & 82.6           & 22.3          & 88.6 & 18.7           & \textbf{50.8} & {\ul 90.3} & 48.4           & 28.9           & \textbf{99.3}  & 17.4           \\
                          & \cellcolor[HTML]{E2EFDA}MYA        & 19.2       & 5.8  & \textbf{100.0} & 18.3          & 0.0 & \textbf{100.0} & 18.5          & 0.9  & \textbf{100.0} & 20.6          & 13.6 & \textbf{100.0} & 32.2          & 88.6 & 38.7           & {\ul 68.5}    & {\ul 90.3} & 97.3           & \textbf{90.4}  & \textbf{99.3}  & 88.0           \\
                          & \cellcolor[HTML]{E2EFDA}DN         & 31.5       & 5.8  & 96.1           & 31.2          & 0.0 & \textbf{100.0} & 31.3          & 0.9  & {\ul 99.3}     & 30.6          & 13.6 & 86.8           & 27.9          & 88.6 & 24.3           & \textbf{67.7} & {\ul 90.3} & 73.0           & {\ul 60.4}     & \textbf{99.3}  & 44.7           \\
                          & \cellcolor[HTML]{FFF2CC}MH         & {\ul 50.0} & 5.8  & \textbf{100.0} & 48.5          & 0.0 & \textbf{100.0} & 48.7          & 0.9  & \textbf{100.0} & \textbf{51.9} & 13.6 & 99.4           & 38.8          & 88.6 & 29.8           & 42.7          & {\ul 90.3} & 32.7           & 35.9           & \textbf{99.3}  & 22.2           \\
                          & \cellcolor[HTML]{FFF2CC}BRVO       & 20.7       & 5.8  & \textbf{100.0} & 19.7          & 0.0 & \textbf{100.0} & 19.8          & 0.9  & \textbf{100.0} & 21.9          & 13.6 & 98.8           & 14.1          & 88.6 & 14.6           & {\ul 66.4}    & {\ul 90.3} & 89.0           & \textbf{85.5}  & \textbf{99.3}  & 79.3           \\
                          & \cellcolor[HTML]{FFF2CC}TSLN       & 11.3       & 5.8  & \textbf{100.0} & 10.7          & 0.0 & \textbf{100.0} & 10.5          & 0.9  & 97.5           & 12.2          & 13.6 & \textbf{100.0} & 18.8          & 88.6 & 30.0           & \textbf{33.3} & {\ul 90.3} & 52.5           & {\ul 30.2}     & \textbf{99.3}  & 20.0           \\
                          & \cellcolor[HTML]{FFF2CC}CSR        & 13.9       & 5.8  & 94.4           & 13.9          & 0.0 & \textbf{100.0} & 14.0          & 0.9  & \textbf{100.0} & 13.8          & 13.6 & 87.0           & 12.9          & 88.6 & 16.7           & \textbf{32.4} & {\ul 90.3} & 42.6           & {\ul 31.4}     & \textbf{99.3}  & 20.4           \\
                          & \cellcolor[HTML]{FFF2CC}CRVO       & 11.8       & 5.8  & \textbf{100.0} & 11.2          & 0.0 & \textbf{100.0} & 11.2          & 0.9  & \textbf{100.0} & 12.7          & 13.6 & \textbf{100.0} & 5.0           & 88.6 & 7.1            & {\ul 54.4}    & {\ul 90.3} & 95.2           & \textbf{85.4}  & \textbf{99.3}  & 83.3           \\
                          & \cellcolor[HTML]{FFF2CC}ODP        & 12.0       & 5.8  & 97.7           & 11.6          & 0.0 & \textbf{100.0} & 11.7          & 0.9  & \textbf{100.0} & 12.1          & 13.6 & 90.9           & 6.5           & 88.6 & 9.1            & \textbf{31.0} & {\ul 90.3} & 45.5           & {\ul 25.0}     & \textbf{99.3}  & 15.9           \\
                          & \cellcolor[HTML]{FFF2CC}ODE        & 20.5       & 5.8  & \textbf{100.0} & 19.5          & 0.0 & \textbf{100.0} & 19.6          & 0.9  & \textbf{100.0} & 21.9          & 13.6 & \textbf{100.0} & 24.6          & 88.6 & 27.2           & {\ul 36.9}    & {\ul 90.3} & 40.7           & \textbf{45.0}  & \textbf{99.3}  & 30.9           \\
                          & \cellcolor[HTML]{FFF2CC}RS         & 17.5       & 5.8  & \textbf{100.0} & 16.7          & 0.0 & \textbf{100.0} & 16.8          & 0.9  & \textbf{100.0} & 18.8          & 13.6 & \textbf{100.0} & 32.7          & 88.6 & 41.8           & {\ul 60.3}    & {\ul 90.3} & 85.1           & \textbf{86.6}  & \textbf{99.3}  & 82.1           \\
                          & \cellcolor[HTML]{FFF2CC}CRS        & 9.7        & 5.8  & \textbf{100.0} & 9.2           & 0.0 & \textbf{100.0} & 9.3           & 0.9  & \textbf{100.0} & 10.5          & 13.6 & \textbf{100.0} & 13.6          & 88.6 & 23.5           & {\ul 44.1}    & {\ul 90.3} & 82.4           & \textbf{78.1}  & \textbf{99.3}  & 73.5           \\
                          & \cellcolor[HTML]{FFF2CC}RPEC       & 5.9        & 5.8  & 87.0           & 6.4           & 0.0 & \textbf{100.0} & 6.5           & 0.9  & \textbf{100.0} & 6.4           & 13.6 & 87.0           & 7.8           & 88.6 & 17.4           & {\ul 27.5}    & {\ul 90.3} & 60.9           & \textbf{44.4}  & \textbf{99.3}  & 34.8           \\ 
\multirow{-17}{*}{RIADD}  & Average                            & 25.6       & 5.8  & 97.6           & 25.2          & 0.0 & \textbf{100.0} & 25.3          & 0.9  & {\ul 99.7}     & 26.3          & 13.6 & 95.2           & 22.1          & 88.6 & 24.0           & {\ul 53.1}    & {\ul 90.3} & 68.8           & \textbf{61.1}  & \textbf{99.3}  & 52.4           \\ \hline \hline
                          & \cellcolor[HTML]{E2EFDA}DR         & 81.9       & 10.5 & 91.5           & 84.8          & 0.0 & \textbf{100.0} & 86.1          & 13.2 & {\ul 99.1}     & 84.3          & 21.1 & 93.4           & 70.8          & 47.4 & 65.1           & \textbf{90.9} & {\ul 94.7} & 84.9           & {\ul 90.7}     & \textbf{100.0} & 83.0           \\
                          & \cellcolor[HTML]{E2EFDA}LOC        & 74.6       & 10.5 & \textbf{100.0} & 72.5          & 0.0 & \textbf{100.0} & 75.2          & 13.2 & \textbf{100.0} & {\ul 76.9}    & 21.1 & \textbf{100.0} & \textbf{83.3} & 47.4 & \textbf{100.0} & 71.6          & {\ul 94.7} & 58.0           & 46.2           & \textbf{100.0} & 30.0           \\
                          & \cellcolor[HTML]{E2EFDA}PM         & 76.1       & 10.5 & \textbf{100.0} & 74.0          & 0.0 & \textbf{100.0} & 76.6          & 13.2 & \textbf{100.0} & 78.3          & 21.1 & \textbf{100.0} & 79.7          & 47.4 & 90.7           & {\ul 98.2}    & {\ul 94.7} & \textbf{100.0} & \textbf{100.0} & \textbf{100.0} & \textbf{100.0} \\
                          & \cellcolor[HTML]{FFF2CC}BRVO       & 72.1       & 10.5 & \textbf{100.0} & 69.8          & 0.0 & \textbf{100.0} & 72.7          & 13.2 & \textbf{100.0} & 74.6          & 21.1 & \textbf{100.0} & 76.9          & 47.4 & 90.9           & {\ul 97.8}    & {\ul 94.7} & \textbf{100.0} & \textbf{100.0} & \textbf{100.0} & \textbf{100.0} \\
                          & \cellcolor[HTML]{FFF2CC}CRVO       & 56.4       & 10.5 & \textbf{100.0} & 53.7          & 0.0 & \textbf{100.0} & 57.1          & 13.2 & \textbf{100.0} & 59.5          & 21.1 & \textbf{100.0} & 66.7          & 47.4 & 95.5           & {\ul 95.7}    & {\ul 94.7} & \textbf{100.0} & \textbf{100.0} & \textbf{100.0} & \textbf{100.0} \\
                          & \cellcolor[HTML]{FFF2CC}RRD        & 77.0       & 10.5 & \textbf{100.0} & 75.0          & 0.0 & \textbf{100.0} & 77.6          & 13.2 & \textbf{100.0} & {\ul 79.2}    & 21.1 & \textbf{100.0} & \textbf{85.1} & 47.4 & \textbf{100.0} & 68.9          & {\ul 94.7} & 54.4           & 77.4           & \textbf{100.0} & 63.2           \\
                          & \cellcolor[HTML]{FFF2CC}MAC        & 81.3       & 10.5 & \textbf{100.0} & 79.6          & 0.0 & \textbf{100.0} & 81.8          & 13.2 & \textbf{100.0} & 83.1          & 21.1 & \textbf{100.0} & 88.1          & 47.4 & \textbf{100.0} & {\ul 98.0}    & {\ul 94.7} & 98.6           & \textbf{99.3}  & \textbf{100.0} & 98.6           \\
                          & \cellcolor[HTML]{FFF2CC}ERM        & 60.5       & 10.5 & \textbf{100.0} & 57.8          & 0.0 & \textbf{100.0} & 61.2          & 13.2 & \textbf{100.0} & 63.4          & 21.1 & \textbf{100.0} & 70.4          & 47.4 & 96.2           & \textbf{90.2} & {\ul 94.7} & 88.5           & {\ul 87.0}     & \textbf{100.0} & 76.9           \\
                          & \cellcolor[HTML]{FFF2CC}MH         & 57.5       & 10.5 & \textbf{100.0} & 54.8          & 0.0 & \textbf{100.0} & 58.2          & 13.2 & \textbf{100.0} & 60.5          & 21.1 & \textbf{100.0} & 65.6          & 47.4 & 91.3           & \textbf{81.0} & {\ul 94.7} & 73.9           & {\ul 75.7}     & \textbf{100.0} & 60.9           \\
                          & \cellcolor[HTML]{FFF2CC}RP         & 56.4       & 10.5 & \textbf{100.0} & 53.7          & 0.0 & \textbf{100.0} & 57.1          & 13.2 & \textbf{100.0} & 59.5          & 21.1 & \textbf{100.0} & 68.8          & 47.4 & \textbf{100.0} & {\ul 95.7}    & {\ul 94.7} & \textbf{100.0} & \textbf{100.0} & \textbf{100.0} & \textbf{100.0} \\
                          & \cellcolor[HTML]{FFF2CC}YWSF       & 63.0       & 10.5 & \textbf{100.0} & 60.4          & 0.0 & \textbf{100.0} & 63.7          & 13.2 & \textbf{100.0} & 65.9          & 21.1 & \textbf{100.0} & 71.1          & 47.4 & 93.1           & {\ul 96.7}    & {\ul 94.7} & \textbf{100.0} & \textbf{100.0} & \textbf{100.0} & \textbf{100.0} \\
                          & \cellcolor[HTML]{FFF2CC}LS         & 54.1       & 10.5 & \textbf{100.0} & 51.3          & 0.0 & \textbf{100.0} & 54.8          & 13.2 & \textbf{100.0} & 57.1          & 21.1 & \textbf{100.0} & 66.7          & 47.4 & \textbf{100.0} & {\ul 95.2}    & {\ul 94.7} & \textbf{100.0} & \textbf{100.0} & \textbf{100.0} & \textbf{100.0} \\
                          & \cellcolor[HTML]{FFF2CC}BF w/o PDR & 87.0       & 10.5 & \textbf{100.0} & 85.3          & 0.0 & 99.1           & {\ul 87.4}    & 13.2 & \textbf{100.0} & \textbf{88.4} & 21.1 & \textbf{100.0} & 86.0          & 47.4 & 88.6           & 29.4          & {\ul 94.7} & 17.5           & 44.9           & \textbf{100.0} & 28.9           \\
                          & \cellcolor[HTML]{FFF2CC}BF w/ PDR  & 72.6       & 10.5 & \textbf{100.0} & 70.3          & 0.0 & \textbf{100.0} & 73.2          & 13.2 & \textbf{100.0} & {\ul 75.0}    & 21.1 & \textbf{100.0} & 70.0          & 47.4 & 77.8           & 65.7          & {\ul 94.7} & 51.1           & \textbf{78.4}  & \textbf{100.0} & 64.4           \\
\multirow{-15}{*}{JSIEC}  & Average                            & 71.2       & 10.5 & 99.3           & 69.3          & 0.0 & \textbf{99.9}  & 72.0          & 13.2 & \textbf{99.9}  & 73.5          & 21.1 & 99.5           & 76.1          & 47.4 & 91.9           & {\ul 83.9}    & {\ul 94.7} & 80.0           & \textbf{85.6}  & \textbf{100.0} & 78.6           \\
\bottomrule
\end{tabular}
}
\end{table}
\noindent\textbf{OCT 2017 is no longer sufficient as a standalone benchmark.}
Figure~\ref{fig:AUROC_abc} (c) presents the results for each abnormal category on OCT test sets.
As shown in the figure, the AUROCs on OCT 2017 are nearly saturated for DRA, particularly for the CNV category, where DRA, PatchCore, and SimpleNet all achieve near-perfect scores.
This saturation indicates that the OCT 2017 dataset is no longer adequate as a standalone benchmark for evaluating retinal anomaly detection methods~\cite{bao2024bmad,cai2024medianomaly,guo2023encoder}, highlighting the importance of our work.

\noindent\textbf{More attention should be paid to threshold-dependent metrics.}
Most anomaly detection methods and benchmarks only take threshold-independent metrics for evaluation~\cite{bao2024bmad,cai2024medianomaly,xie2024iad}, which limits the evaluation of their readiness for real-world deployment.
Benchmarks and analyses for threshold-dependent metrics are critical to draw more attention to the evaluation tailored for clinical practice.
As shown in Table~\ref{tab: RIADD} and~\ref{tab: OCT}, the F1 scores, specificities, and sensitivities reveal that while the proposed NFM-DRA achieves the highest average F1 scores, no single method consistently outperforms others across different datasets and anomaly categories.
Moreover, the performance of threshold-dependent metrics exhibits more significant fluctuations compared to threshold-agnostic metrics.
We hope this benchmark will inspire a rethinking of evaluation protocols for practical anomaly detection.
\subsection{Implementation}
We conduct benchmarking following the original papers and official codebases based on PyTorch~\cite{paszke2019pytorch}.
Grid search is employed to identify the optimal learning rate and number of epochs around the default settings.
The best-performing checkpoints on the validation set are evaluated on the test set.
All experiments are performed on a single NVIDIA RTX 3090 GPU.
\begin{table}[!t]
\caption{Threshold-dependent results (\%) on OCT test sets, presenting F1 scores, specificity and sensitivity.
\colorbox{seencolor}{Seen} and \colorbox{unseencolor}{unseen} anomaly categories are highlighted to show whether they are included in the training set.
The \textbf{best} results are in bold and the \uline{second-best} ones are underlined.}
\label{tab: OCT}
\centering
\resizebox{\linewidth}{!}{
\begin{tabular}{c|c|ccc|ccc|ccc|ccc|ccc|ccc|ccc}
\toprule
                                                                          &                                & \multicolumn{3}{c|}{SoftPatch}         & \multicolumn{3}{c|}{EDC}               & \multicolumn{3}{c|}{SimpleNet}                           & \multicolumn{3}{c|}{PatchCore}                     & \multicolumn{3}{c|}{DDAD-ASR}         & \multicolumn{3}{c|}{DRA}                                 & \multicolumn{3}{c}{NFM-DRA (Ours)}                     \\ \cline{3-23} 
\multirow{-2}{*}{Dataset}                                                 & \multirow{-2}{*}{Normal vs}    & F1   & SPC  & \multicolumn{1}{c|}{SEN} & F1   & SPC  & \multicolumn{1}{c|}{SEN} & F1            & SPC           & \multicolumn{1}{c|}{SEN} & F1         & SPC        & \multicolumn{1}{c|}{SEN} & F1   & SPC & \multicolumn{1}{c|}{SEN} & F1            & SPC           & \multicolumn{1}{c|}{SEN} & F1            & SPC            & SEN            \\ \hline
                                                                          & \cellcolor[HTML]{E2EFDA}CNV    & 75.0 & 34.0 & 99.6                     & 71.9 & 22.0 & \textbf{100.0}           & 89.0          & 75.2          & \textbf{100.0}           & 91.6       & 81.6       & \textbf{100.0}           & 68.1 & 6.4 & \textbf{100.0}           & \textbf{99.2} & \textbf{98.4} & \textbf{100.0}           & {\ul 98.8}    & {\ul 97.6}     & \textbf{100.0} \\
                                                                          & \cellcolor[HTML]{E2EFDA}DME    & 75.2 & 34.0 & \textbf{100.0}           & 71.9 & 22.0 & \textbf{100.0}           & 88.8          & 75.2          & 99.6                     & 91.6       & 81.6       & \textbf{100.0}           & 67.9 & 6.4 & 99.6                     & \textbf{99.2} & \textbf{98.4} & \textbf{100.0}           & {\ul 98.8}    & {\ul 97.6}     & \textbf{100.0} \\
                                                                          & \cellcolor[HTML]{E2EFDA}DN & 74.4 & 34.0 & 98.4                     & 70.3 & 22.0 & 96.4                     & 87.4          & 75.2          & 96.8                     & 91.0       & 81.6       & 98.8                     & 67.6 & 6.4 & 98.8                     & \textbf{99.2} & \textbf{98.4} & \textbf{100.0}           & {\ul 98.6}    & {\ul 97.6}     & {\ul 99.6}     \\
\multirow{-4}{*}{\begin{tabular}[c]{@{}c@{}}OCT\\      2017\end{tabular}} & Average                        & 78.6 & 34.0 & 99.3                     & 75.5 & 22.0 & 98.8                     & 90.1          & 75.2          & 98.8                     & 92.7       & 81.6       & 99.6                     & 72.5 & 6.4 & 99.5                     & \textbf{99.3} & \textbf{98.4} & \textbf{100.0}           & {\ul 98.9}    & {\ul 97.6}     & {\ul 99.9}     \\ \hline \hline
                                                                          & \cellcolor[HTML]{E2EFDA}AMD    & 90.4 & 33.7 & 97.2                     & 88.1 & 0.0  & \textbf{100.0}           & 84.8          & \textbf{83.4} & 76.8                     & {\ul 92.2} & 45.2       & 98.1                     & 88.1 & 0.0 & \textbf{100.0}           & 89.6          & 15.1          & 99.8                     & \textbf{94.7} & {\ul 61.4}     & 99.2           \\
                                                                          & \cellcolor[HTML]{E2EFDA}DME    & 55.8 & 33.7 & 96.6                     & 47.0 & 0.0  & \textbf{100.0}           & \textbf{72.2} & \textbf{83.4} & 77.6                     & 60.3       & 45.2       & 96.6                     & 47.0 & 0.0 & \textbf{100.0}           & 51.0          & 15.1          & \textbf{100.0}           & {\ul 69.0}    & {\ul 61.4}     & 98.6           \\
                                                                          & \cellcolor[HTML]{FFF2CC}ERM    & 54.9 & 33.7 & 91.6                     & 48.3 & 0.0  & \textbf{100.0}           & 58.6          & \textbf{83.4} & 56.1                     & {\ul 59.0} & 45.2       & 91.0                     & 48.3 & 0.0 & \textbf{100.0}           & 51.4          & 15.1          & 97.4                     & \textbf{65.9} & {\ul 61.4}     & 89.7           \\
                                                                          & \cellcolor[HTML]{FFF2CC}RAO    & 16.7 & 33.7 & \textbf{100.0}           & 11.7 & 0.0  & \textbf{100.0}           & \textbf{41.2} & \textbf{83.4} & 90.9                     & 19.5       & 45.2       & \textbf{100.0}           & 11.7 & 0.0 & \textbf{100.0}           & 12.9          & 15.1          & 95.5                     & {\ul 24.6}    & {\ul 61.4}     & 95.5           \\
                                                                          & \cellcolor[HTML]{FFF2CC}RVO    & 46.4 & 33.7 & 96.0                     & 37.8 & 0.0  & \textbf{100.0}           & \textbf{66.7} & \textbf{83.4} & 77.2                     & 49.9       & 45.2       & 93.1                     & 37.8 & 0.0 & \textbf{100.0}           & 41.7          & 15.1          & \textbf{100.0}           & {\ul 58.2}    & {\ul 61.4}     & 93.1           \\
                                                                          & \cellcolor[HTML]{FFF2CC}VID    & 40.0 & 33.7 & 97.4                     & 31.4 & 0.0  & \textbf{100.0}           & \textbf{60.6} & \textbf{83.4} & 75.0                     & 44.6       & 45.2       & 97.4                     & 31.4 & 0.0 & \textbf{100.0}           & 35.0          & 15.1          & \textbf{100.0}           & {\ul 53.2}    & {\ul 61.4}     & 97.4           \\
\multirow{-7}{*}{OCTDL}                                                   & Average                        & 56.6 & 33.7 & 96.5                     & 50.8 & 0.0  & \textbf{100.0}           & \textbf{66.9} & \textbf{83.4} & 75.5                     & 59.8       & 45.2       & 96.2                     & 50.8 & 0.0 & \textbf{100.0}           & 53.4          & 15.1          & 98.9                     & {\ul 65.8}    & {\ul 61.4}     & 95.9           \\ \hline \hline
                                                                          & \cellcolor[HTML]{E2EFDA}AMD    & 63.0 & 72.8 & 92.7                     & 34.8 & 0.0  & \textbf{100.0}           & 42.6          & 28.2          & \textbf{100.0}           & 82.9       & {\ul 95.1} & 83.6                     & 34.8 & 0.0 & \textbf{100.0}           & {\ul 90.8}    & {\ul 95.1}    & 98.2                     & \textbf{96.2} & \textbf{100.0} & 92.7           \\
                                                                          & \cellcolor[HTML]{E2EFDA}DR     & 77.4 & 72.8 & 96.3                     & 51.0 & 0.0  & \textbf{100.0}           & 56.7          & 28.2          & 94.4                     & 89.6       & {\ul 95.1} & 88.8                     & 51.0 & 0.0 & \textbf{100.0}           & {\ul 93.6}    & {\ul 95.1}    & 96.3                     & \textbf{96.1} & \textbf{100.0} & 92.5           \\
                                                                          & \cellcolor[HTML]{FFF2CC}CSR    & 76.6 & 72.8 & 96.1                     & 49.8 & 0.0  & \textbf{100.0}           & 55.5          & 28.2          & 94.1                     & 89.7       & {\ul 95.1} & 89.2                     & 49.8 & 0.0 & \textbf{100.0}           & {\ul 93.3}    & {\ul 95.1}    & 96.1                     & \textbf{96.4} & \textbf{100.0} & 93.1           \\
                                                                          & \cellcolor[HTML]{FFF2CC}MH     & 78.5 & 72.8 & \textbf{100.0}           & 49.8 & 0.0  & \textbf{100.0}           & 56.7          & 28.2          & 97.1                     & 93.8       & {\ul 95.1} & 97.1                     & 49.8 & 0.0 & \textbf{100.0}           & {\ul 94.3}    & {\ul 95.1}    & 98.0                     & \textbf{98.0} & \textbf{100.0} & 96.1           \\
\multirow{-5}{*}{OCTID}                                                   & Average                        & 77.3 & 72.8 & 96.4                     & 52.7 & 0.0  & \textbf{100.0}           & 58.6          & 28.2          & 96.3                     & 89.9       & {\ul 95.1} & 89.8                     & 52.7 & 0.0 & \textbf{100.0}           & {\ul 93.8}    & {\ul 95.1}    & 97.1                     & \textbf{96.7} & \textbf{100.0} & 93.6          \\
\bottomrule
\end{tabular}
}
\end{table}
\section{Conclusion}
In this paper, we establish a systematic evaluation framework based on seven public datasets.
We evaluate anomaly detection methods under different levels of supervision.
DRA, a fully supervised approach, achieves the best overall performance, while PatchCore, a memory bank-based one-class supervised method, demonstrates greater robustness for unseen anomalies.
Based on these findings, we propose NFM-DRA to enhance DRA with a novel feature memory, which outperforms all benchmarked methods and exhibits improved robustness.
We hope the proposed benchmark will inspire rigorous evaluation of retinal anomaly detection and foster impactful research in this field.
%
%
%
\bibliographystyle{splncs04}
\bibliography{mybibliography}
\end{document}